\documentclass[10pt,twocolumn,letterpaper]{article}
\usepackage{cvpr}
\usepackage{times}
\usepackage{color}
\usepackage{epsfig}
\usepackage{graphicx}
\usepackage{amsmath}
\usepackage{amssymb}
\usepackage{amssymb}
\usepackage{booktabs}
\usepackage{mathrsfs}
\usepackage{enumitem}
\usepackage[utf8]{inputenc}
\usepackage{multirow}
\usepackage[pagebackref=true,breaklinks=true,letterpaper=true,colorlinks,bookmarks=false]{hyperref}

\cvprfinalcopy 

\def\cvprPaperID{3126} 

\ifcvprfinal\fi
\begin{document}
	
	\title{Deep Learning Human Mind for Automated Visual Classification}
	\author{C. Spampinato, S. Palazzo, I. Kavasidis, D. Giordano\\
		Department of Electrical, Electronics and Computer Engineering - PeRCeiVe Lab\\
		Viale Andrea Doria, 6 - 95125 Catania\\
		{\tt\small http://perceive.dieei.unict.it}
		\and
		N. Souly, M. Shah\\
		Center for Research in Computer Vision -- University of Central Florida\\
		4328 Scorpius St.,	HEC 245D Orlando, FL 32816-2365\\
		{\tt\small http://crcv.ucf.edu/}
	}
	
	\maketitle

\begin{abstract}
  What if we could effectively read the mind and transfer human visual capabilities to computer vision methods?  In this paper, we aim at addressing this question by developing the {\it first visual object classifier driven by human brain signals}.
  In particular, we employ EEG data evoked by visual object stimuli combined with Recurrent Neural Networks (RNN) to learn a discriminative brain activity manifold of visual categories in a reading the mind effort. Afterward, we transfer the learned capabilities to machines by training a Convolutional Neural Network (CNN)--based regressor to project images onto the learned manifold, thus allowing machines to employ human brain--based features for automated visual classification.
  We use a 128-channel EEG with active electrodes to record brain activity of several subjects while looking at images of 40 ImageNet object classes. The proposed RNN-based approach for discriminating object classes using brain signals reaches an average accuracy of about 83\%, which greatly outperforms existing methods attempting to learn EEG visual object representations. As for automated object categorization, our human brain--driven approach obtains competitive performance, comparable to those achieved by powerful CNN models and it is also able to generalize over different visual datasets. 
\end{abstract}

\section{Introduction}
\begin{center}
	\begin{figure*}
		\centering
		\includegraphics[width=0.9\textwidth]{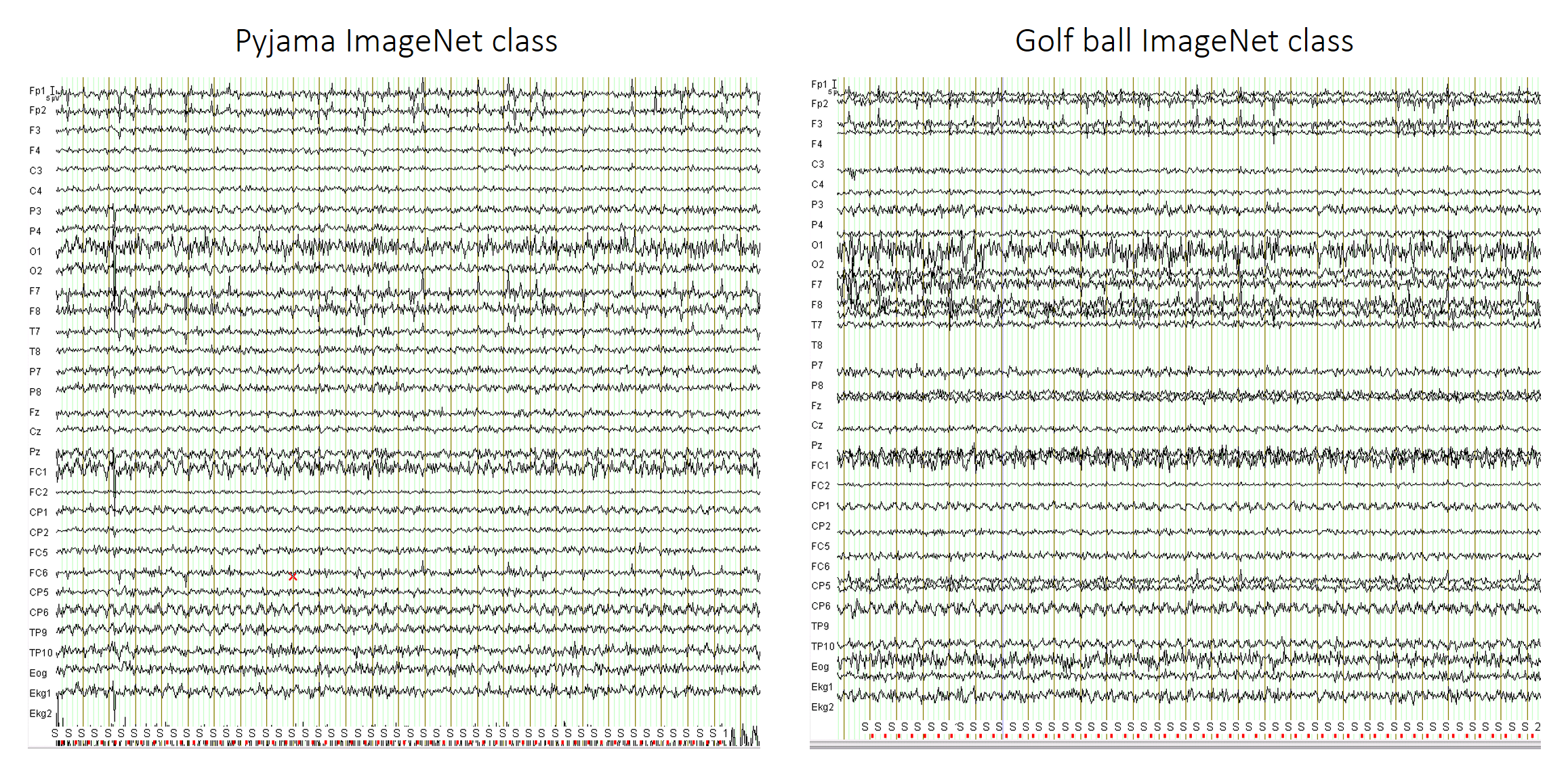}
		\caption{Examples of brain signals evoked by visual stimuli of two different ImageNet object classes.}
		\label{fig:eeg_signals}
	\end{figure*}
\end{center}
Humans show excellent performance, still unreachable by machines, in interpreting visual scenes. Despite the recent rediscovery of Convolutional Neural Networks has led to a significant performance improvement in automated visual classification, their generalization capabilities are not at the human level, since they learn a discriminative feature space, which strictly depends on the employed training dataset rather than on more general principles. More specifically, the first-layer features of a CNN appear to be generalizable across different datasets, as they are similar to Gabor filters and color blobs, while the last-layer features are very specific to a particular dataset or task. 
In humans, instead, the process behind visual object recognition stands at the interface between perception, i.e., how objects appear visually in terms of shape, colors, etc. (all features that can be modeled with first CNN layers) and conception, which involves higher cognitive processes that have never been exploited.
Several cognitive neuroscience studies~\cite{pmid10777794,pmid18829969,pmid17643089} have investigated which parts of visual cortex and brain are responsible for such cognitive processes, but, so far, there is no clear solution. Of course, this reflects on the difficulties of cognition-based automated methods to perform visual tasks.\\[0 pt]
We argue that one possible solution is to act in a reverse engineering manner, i.e., by analyzing human brain activity -- recorded through neurophysiology (EEG/MEG) and neuroimaging techniques (e.g., fMRI) -- to identify the feature space employed by humans for visual classification. 
In relation to this, it is has been acknowledged that brain activity recordings contain information about visual object categories~\cite{pmid20302949,pmid22983495,export:64271,pmid21920851,pmid23908380,Kaneshiro2015,Simanova2010}.
Understanding EEG data evoked by specific stimuli has been the goal of brain computer interfaces (BCI) research for years. Nevertheless, BCIs aim mainly at classifying or detecting specific brain signals to allow direct-actuated control of machines for disabled people. 
In this paper, we want to take a great leap forward with respect to classic BCI approaches, i.e., we aim at exploring a new and direct form of human involvement (a new vision of the ``human-based computation'' strategy) for automated visual classification.
The underlying idea is to learn a brain signal discriminative manifold of visual categories by classifying EEG signals - {\it reading the mind} - and then to project images into such manifold to allow machines to perform automatic  visual categorization - {\it transfer human visual capabilities to machines}.
The impact of decoding object category--related EEG signals for inclusion into computer vision methods is tremendous.
First, identifying EEG-based discriminative features for visual categorization might provide meaningful insight about the human visual perception systems. As a consequence, it will greatly advance performance of BCI-based applications as well as enable a new form of brain-based image labeling.
Second, effectively projecting images into a new biologically based manifold will change radically the way object classifiers are developed (mainly in terms of feature extraction).  
Thus, the contribution of this paper is threefold:
\begin{itemize}
\item We propose a deep learning approach to classify EEG data evoked by visual object stimuli outperforming state-of-the-art methods both in the number of tackled object classes and in classification accuracy.
\item We propose the {\it first computer vision approach driven by brain signals}, i.e.,  the first automated classification approach employing visual descriptors extracted directly from human neural processes involved in visual scene analysis.
\item We will publicly release the largest EEG dataset for visual object analysis, with related source code and trained models.
\end{itemize}

\section{Related Work}
\begin{figure*}
	\centering
	\includegraphics[width=1\textwidth]{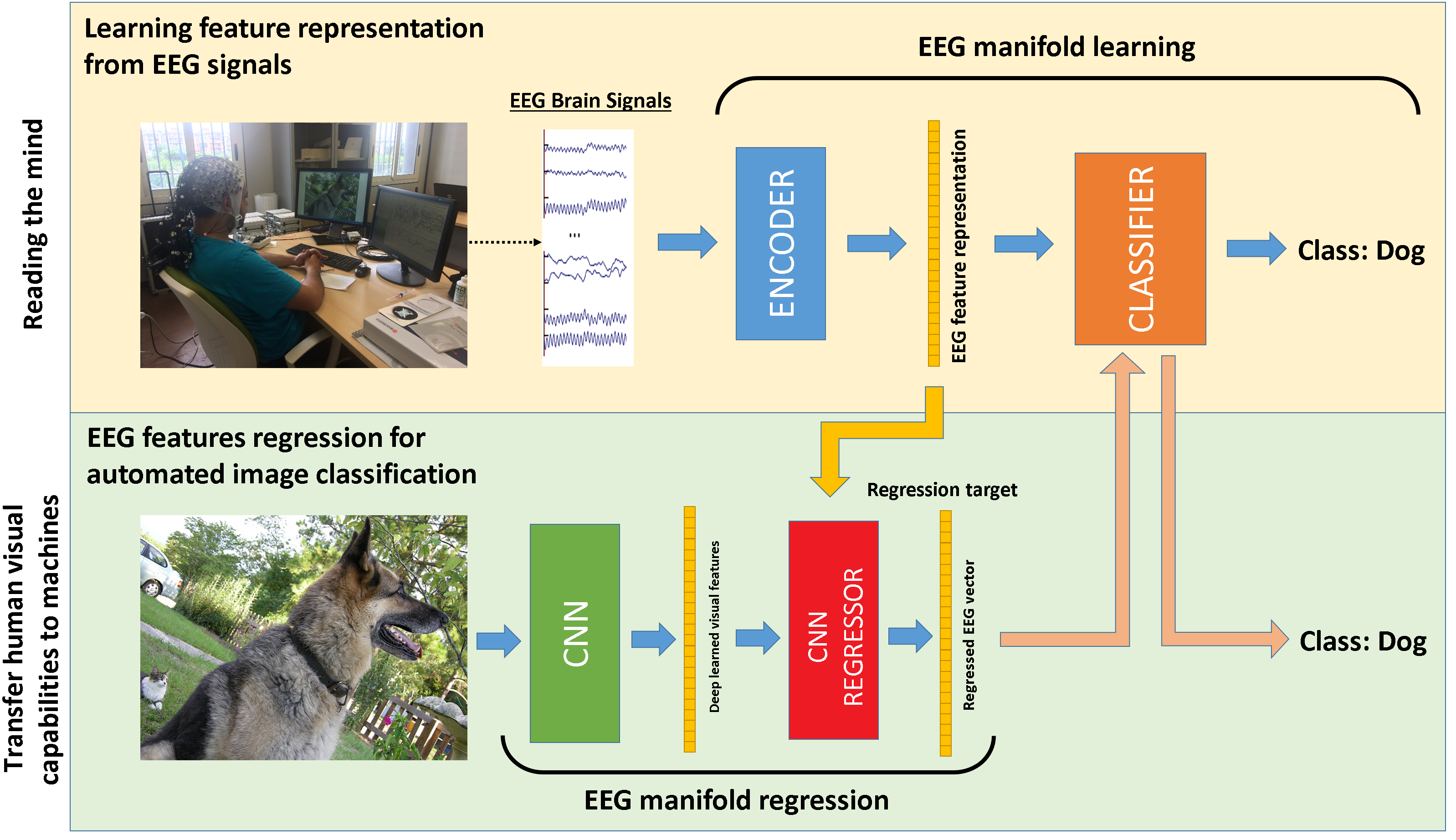}
	\caption{Overview of the proposed approach. Top: a low-dimensional representation for temporal EEG signals recorded while users looked at images is learned by the \textit{encoder} module; the computed EEG features are employed to train an image classifier. Bottom: a CNN is trained to estimate EEG features directly from images; then, the classifier trained in the previous stage can be used for automated classification without the need of EEG data for new images.}
	\label{fig:method_overview}
\end{figure*}

The idea of reading the mind of people while performing specific tasks has been long investigated, especially for building brain-computer interfaces. Most of BCI studies have mainly performed binary EEG-data classification, i.e., presence of absence of a specific pattern, e.g., in~\cite{5492691} for P300 detection or in~\cite{pmid19837629} for seizure detection.

Recently, thanks to deep learning, other works have attempted to investigate how to model more complex cognitive events (e.g., cognitive load, audio stimuli, etc.) from brain signals. For example, in~\cite{BashivanRYC15}, a combination of recurrent and convolutional neural networks was proposed to learn EEG representations for cognitive load classification task (reported classification accuracy is of about 90\% over four cognitive load levels). In~\cite{stober}, a similar approach, using only CNNs, is proposed to learn to classify EEG-recordings evoked by audio music with an accuracy of 28\% over 12 songs. These methods have proved the potential of using brain signals and deep learning for classification, but they tackle a small number of classification categories (maximum twelve in~\cite{stober}), and none of them are related to visual scene understanding.

A number of cognitive neuroscience studies have demonstrated (by identifying specific regions of visual cortex) that up to a dozen of object categories can be decoded in event-related potential (ERP)
amplitudes recorded through EEG~\cite{pmid22983495,pmid21920851,Simanova2010}.
However, such scientific evidence has not been deeply exploited to build visual stimuli--evoked EEG classifiers. Indeed, a very limited number of methods have been developed~\cite{Shamlo2008,Kapoor2008,Stewart2014,Kaneshiro2015} (none of them using deep learning) to address the problem of decoding visual object--related EEG data, and most of these methods were mainly devised for binary classification (e.g., presence or absence of a given object class). One of the most recent and comprehensive methods was proposed by Kaneshiro {\it et al.} in \cite{Kaneshiro2015}, who trained a classifier able to distinguish EEG brain signals evoked by twelve different object classes, with an accuracy of about 29\% and that represents, so far, the state-of-art performance.

In this paper, we explore not only the capabilities of deep learning in modeling visual stimuli--evoked EEG with more object classes than state-of-the-art methods, but we also investigate how to project images into an EEG-based manifold in order to allow machines to interpret visual scenes automatically using features extracted according to human brain processes. This, to the best of our knowledge, has not been done before.

\begin{center}
	\begin{figure*}
		\centering
		\includegraphics[width=0.8\textwidth]{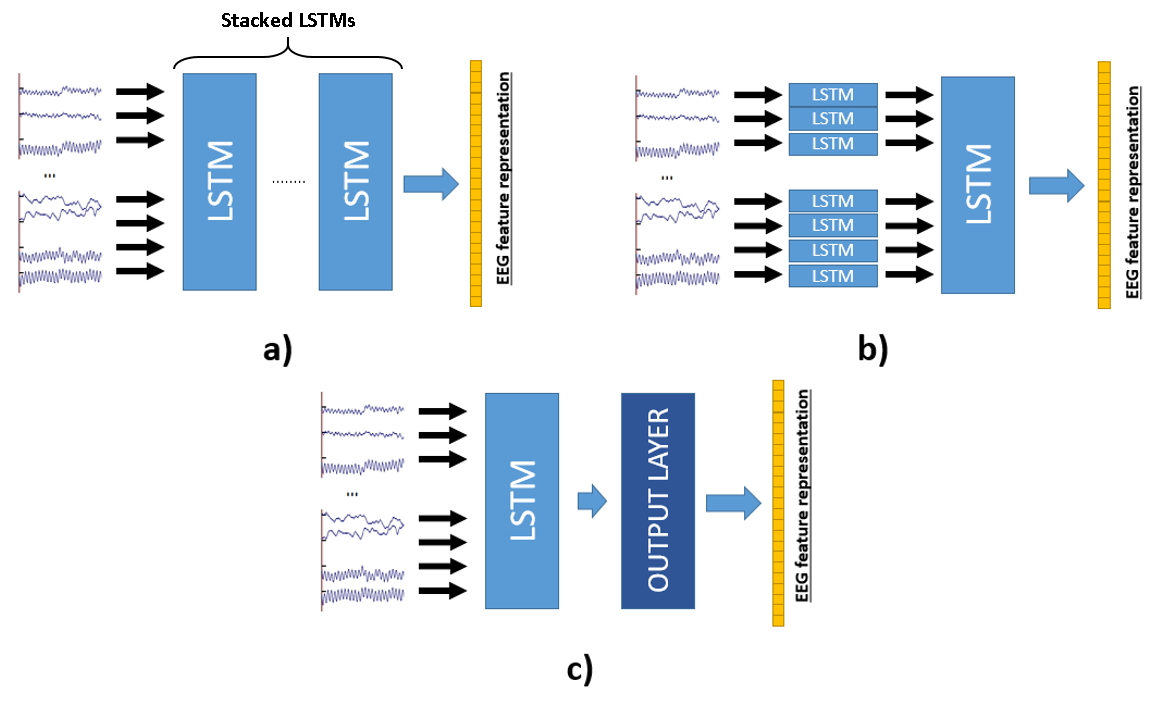}
		\caption{Tested encoder architectures: \textit{a)} common LSTM; \textit{b)} channel LSTM + common LSTM; \textit{c)} common LSTM + output layer.}
		\label{fig:rnn_model}
	\end{figure*}
\end{center}

\section{Method}
The work described in this paper relies on three key intuitions:
\begin{itemize}
 \item EEG signals recorded while a subject looks at an image convey feature-level and cognitive-level information about the image content (a qualitative difference between EEG signals evoked, on one subject, by visual stimuli of two different object classes is shown in Fig.~\ref{fig:eeg_signals}).
 \item A low-dimensional manifold within the multi-dimensional and temporally-varying EEG signals exists and can be extracted to obtain a 1D representation which we refer to as \textit{EEG features}.
 \item {\it EEG features} are assumed to mainly encode visual data, thus it is possible to extract the corresponding image descriptors for automated classification.
\end{itemize}
These three ideas provide the design basis for the overall two-stage image classification architecture proposed in this work and shown in Fig.~\ref{fig:method_overview}.
The first stage of our approach - the {\it reading the mind} phase - aims at identifying a low-dimensional manifold within the two-dimensional (channels and time) EEG space, such that the representation within that manifold is discriminant over object classes. In order to learn this representation, we employed EEG data recorded while users looked at images on a screen. Then, we trained an \textit{encoder} network (implemented through recurrent neural networks -- RNNs -- for temporal analysis) to extract \textit{EEG features} from raw EEG signals; the training process is supervised by the class of the images for which each input EEG sequences were recorded, and a classifier for EEG features is jointly trained in the process.

Of course, it is unreasonable to assume the availability of  EEG data for each image to be classified. Therefore, the second stage of the method aims at extracting {\it EEG features} directly from images - the {\it transfer human visual capabilities to machines} phase - by learning a mapping from CNN deep visual descriptors to EEG features (learned through RNN encoder). After that, new images can be classified by simply estimating their {\it EEG features} through the trained CNN-based regressor and employ the stage-one classifier to predict the corresponding image class.

	\begin{table}
		\centering
		\begin{tabular}{lr}
			\toprule
			Number of classes &  40 \\
			Number of images per class &  50 \\
			Total number of images &  2000 \\
			Visualization order &  Sequential \\
			Time for each image & 0.5 s \\
			Pause time between classes  & 10 s \\
			Number of sessions & 4 \\
			Session running time & 350 s\\
			Total running time & 1400 s\\
			\bottomrule
		\end{tabular}
		\caption{The parameters of the experimental protocol.}
		\label{tab:paradigms}
	\end{table}

\subsection{EEG data acquisition}
\label{sec:eeg_data}
Six subjects (five male and one female) were shown visual stimuli of objects while EEG data was recorded. All subjects were homogeneous in terms of age, education level and cultural background and were evaluated by a professional physicist in order to exclude possible conditions (e.g., diseases) interfering with the acquisition process.

The dataset used for visual stimuli was a subset of ImageNet~\cite{ILSVRC15}, containing 40 classes of easily recognizable objects\footnote{ImageNet classes used: dog, cat, butterfly, sorrel, capuchin, elephant, panda, fish, airliner, broom, canoe, phone, mug, convertible, computer, watch, guitar, locomotive, espresso, chair, golf, piano, iron, jack, mailbag, missile, mitten, bike, tent, pajama, parachute, pool, radio, camera, gun, shoe, banana, pizza, daisy and bolete (fungus)}.  
During the experiment, 2,000 images (50 from each class) were shown in bursts for 0.5 seconds each. A burst lasts for 25 seconds, followed by a 10-second pause where a black image was shown for a total running time of 1,400 seconds (23 minutes and 20 seconds). 
A summary of the adopted experimental paradigm is shown in Table~\ref{tab:paradigms}.

The experiments were conducted using a 128-channel cap with active, low-impedance electrodes (actiCAP 128Ch\footnote{http://www.brainproducts.com/}).
Brainvision\footnote{http://www.brainvision.com/} DAQs and amplifiers were used for the EEG data acquisition. Sampling frequency and data resolution were set, respectively, to 1000 Hz and 16 bits.

A notch filter (49-51 Hz) and a second-order band-pass Butterworth filter (low cut-off frequency 14 Hz, high cut-off frequency 71 Hz) were set up so that the recorded signal included the Beta (15-31 Hz) and Gamma (32-70 Hz) bands, as they convey information about the cognitive processes involved in the visual perception ~\cite{niedermeyer2005electroencephalography}.

From each recorded EEG sequence, the first 40 samples (40 ms) for each image were discarded in order to exclude any possible interference from the previously shown image (i.e., to permit the stimulus to propagate from the retina through the optical tract to the primary visual cortex~\cite{heckenlively2006principles}). The following 440 samples (440 ms) were used for the experiments.
Data value distribution was centered around zero, thus non-linear quantization was applied.  By using the protocol described above we acquired 12,000 (2,000 images for 6 subjects) 128-channel EEG sequences.
In the following descriptions, we will refer to a generic input EEG sequence as $s(c,t)$, where $c$ (from 1 to 128) indexes a channel and $t$ (from 1 to 440) indexes a sample in time. We will also use the symbol $(\cdot)$ to indicate ``all values'', so $s(\cdot,t)$ represents the vector of all channel values at time $t$, and $s(c,\cdot)$ represents the whole set of time samples for channel $c$.

\begin{center}
	\begin{figure*}
		\centering
		\includegraphics[width=0.9\textwidth]{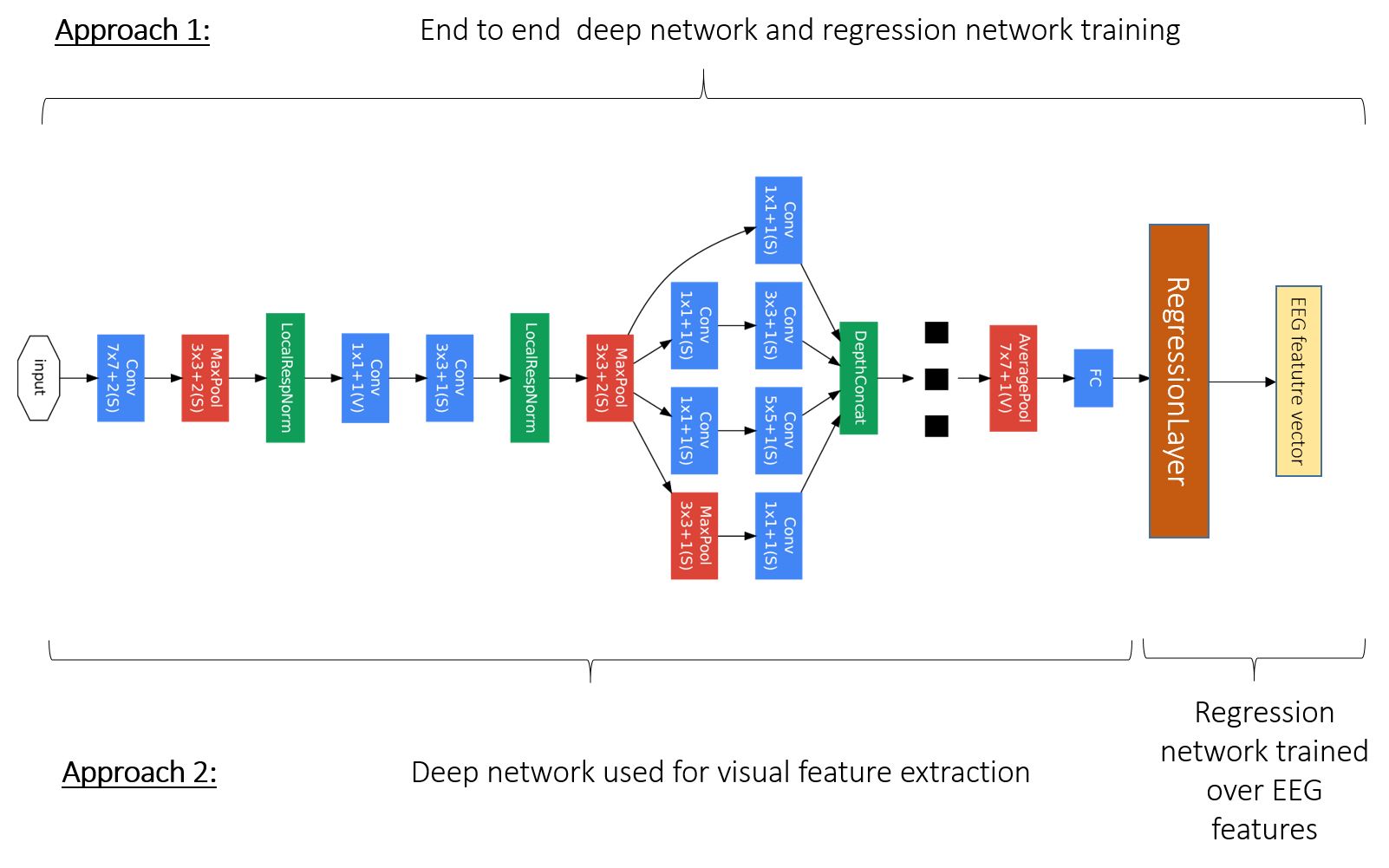}
		\caption{Tested CNN-based regressors. \textit{Approach 1}: we stacked a regression layer to a common deep network and then trained, end to end, the resulting module; \textit{Approach 2}: We extracted deep features using a common off-the-shelf deep network and then train separately the regressor}
		\label{fig:cnn_regressor_model}
	\end{figure*}
\end{center}
\subsection{Learning EEG manifold} 
\label{sec:rnn}
The first analysis aims at translating an input multi-channel temporal EEG sequence into a low dimensional feature vector summarizing the relevant content of the input sequence. Previous approaches~\cite{Kaneshiro2015,Stewart2014} simply concatenate time sequences from multiple channels into a single feature vector, ignoring temporal dynamics, which, instead, contains fundamental information for EEG activity understanding~\cite{Kaneshiro2015}. In order to include such dynamics in our representation, we employ LSTM recurrent neural networks~\cite{Hochreiter:1997:LSM:1246443.1246450} because of their capability to track long-term dependencies in the input data.	
The top half of Fig.~\ref{fig:method_overview} shows the general architecture of our EEG manifold representation model. The EEG multi-channel temporal signals, preprocessed as described in Sect.~\ref{sec:eeg_data}, are provided as input to an \textit{encoder} module, which processes the whole time sequence and outputs an \textit{EEG feature vector} as a compact representation of the input. Ideally, if an input sequence consists of the EEG signals recorded while looking at an image, our objective is to have the resulting output vector encode relevant brain activity information for discriminating different image classes. The encoder network is trained by adding, at its output, a classification module (in all our experiments, it will be a \textit{softmax} layer), and using gradient descent to learn the whole model's parameters end-to-end.
In our experiments, we tested several configurations of the encoder network:
\begin{itemize}
 \item \textit{Common LSTM} (Fig.~\ref{fig:rnn_model}\textit{a}): the encoder network is made up of a stack of LSTM layers. At each time step $t$, the first layer takes the input $s(\cdot,t)$ (in this sense, ``common'' means that all EEG channels are initially fed into the same LSTM layer); if other LSTM layers are present, the output of the first layer (which may have a different size than the original input) is provided as input to the second layer and so on. The output of the deepest LSTM layer at the last time step is used as the EEG feature representation for the whole input sequence.
 \item \textit{Channel LSTM + Common LSTM} (Fig.~\ref{fig:rnn_model}\textit{b}): the first encoding layer consists of several LSTMs, each connected to only one input channel: for example, the first LSTM processes input data $s(1,\cdot)$, the second LSTM processes $s(2,\cdot)$, and so on. In this way, the output of each ``channel LSTM'' is a summary of a single channel's data. The second encoding layer then performs inter-channel analysis, by receiving as input the concatenated output vectors of all channel LSTMs. As above, the output of the deepest LSTM at the last time step is used as the encoder's output vector.
 \item \textit{Common LSTM + output layer} (Fig.~\ref{fig:rnn_model}\textit{c}): similar to the \textit{common LSTM} architecture, but an additional output layer (linear combinations of input, followed by ReLU nonlinearity) is added after the LSTM, in order to increase model capacity at little computational expenses (if compared to the two-layer common LSTM architecture). In this case, the encoded feature vector is the output of the final layer.
\end{itemize}
Encoder and classifier training is performed through gradient descent by providing the class label associated to the image shown while each EEG sequence was recorded. After training, the encoder  can be used to generate EEG features from an input EEG sequences, while the classification network will be used to predict the image class for an input EEG feature representation, which can be computed from either EEG signals or images, as described in the next section.

\subsection{CNN-based Regression on EEG manifold for Visual Classification}
In order to employ the RNN learned feature representation for general images, it is necessary to bypass the EEG recording stage and extract features directly from the image, which should be possible by our assumption that the learned EEG features reflect the image content which evoked the original EEG signals.

We employed two CNN-based approaches (see Fig.~\ref{fig:cnn_regressor_model}) to extract EEG features (or, at least, a close approximation) from an input image:
\begin{itemize}[nolistsep,noitemsep]
 \item {\it Approach 1: End to end training.} The first approach is to train a CNN to map images to corresponding EEG feature vectors. Typically, the first layers of CNN attempt to learn the general (global) features of the images, which are common between many tasks, thus we initialize the weights of these layers using pre-trained models, and then learn the weights of last layers from scratch in an end to end setting.  In particular, we used the pre-trained AlexNet CNN~\cite{krizhevsky2012imagenet}, and modified it by replacing the softmax classification layer with a regression layer (containing as many neurons as the dimensionality of the EEG feature vectors), using Euclidean loss as objective function.
 \item {\it Approach 2: Deep feature extraction followed by regressor training.} The second approach consists of extracting image features using pre-trained CNN models and then employ regression methods to map image features to EEG feature vectors. We used our fine-tuned AlexNet~\cite{krizhevsky2012imagenet}, GoogleNet~\cite{szegedy2015going} and VGG~\cite{simonyan2014very} as feature extractors by reading the output of the last fully-connected layer, and then applied several regression methods (namely, k-NN regression, ridge regression, random forest regression) to obtain the predicted feature vectors.
 \end{itemize}

We opted to fine-tune only AlexNet, instead of GoogleNet~\cite{szegedy2015going} and VGG~\cite{simonyan2014very}, because these two CNNs contain more convolutional layers and, as such, they were more prone to overfitting given the relatively small dataset size.
The resulting CNN-based regressor is able to extract brain-learned features from any input image for futher classification by the softmax layer trained during EEG feature learning.

\section{Performance Analysis}
Performance analysis is split into three parts since our method consists of: 1) learning visual stimuli--evoked EEG data using RNN (implemented in Torch\footnote{http://torch.ch/}); 2) CNN-based regression to map images to RNN-learned EEG-based features (implemented in Caffe\footnote{http://caffe.berkeleyvision.org/}); 3) the combination of the above two steps to implement automated visual classifiers.

\begin{table*}[]
	\centering
	\begin{tabular}{cccc}
		\toprule
		\textbf{Model}                 & \textbf{Details} & \textbf{Max VA} & \textbf{TA at max VA} \\
		\midrule
		\multirow{3}{*}{Common} & 64 common                     & 74.4\%                                & 73.9\%                              \\
		& 128 common                    & 77.3\%                                & 74.1\%                              \\
		& 64,64 common                  & 75.9\%                                & 72.5\%                              \\
		& 128,64 common                 & 79.1\%                                & 76.8\%                              \\
		& 128,128 common                & 79.7\%                                & 78.0\%                              \\
		\midrule
		\multirow{2}{*}{Channel + Common}           & 5 channel, 32 common          & 75.7\%                                & 72.9\%                              \\
		& 5 channel, 64 common          & 74.3\%                                & 71.2\%                              \\
		\midrule
		\multirow{2}{*}{Common + output}           & 128 common, 64 output        & 81.6\%                                & 78.7\%                              \\
		& \textbf{128 common, 128 output}       & \textbf{85.4\%}              & \textbf{82.9\%}                              \\
		\bottomrule
	\end{tabular}
	\caption{Maximum validation accuracy (``Max VA'') and corresponding test accuracy (``TA at max VA'') for different configurations of the three RNN architectures shown in Sect.~\ref{sec:rnn}. The model yielding the best validation results is in bold.}
	\label{tab:results_rnn}
\end{table*}

\subsection{Learning visual stimuli--evoked EEG representations}

We first tested the three architectures reported in Sect.~\ref{sec:rnn} using our EEG dataset. Our dataset was split into training, validation and test sets, with respective fractions 80\% (1600 images), 10\% (200), 10\% (200). We ensured that the signals generated by all participants for a single image are all included in a single split. All model architecture choices were taken only based on the results on the validation split, making the test split a reliable and ``uncontaminated'' quality indicator for final evaluations. The overall number of EEG sequences used for training the RNN encoder was 12,000.

Existing works, such as~\cite{Subasi2010,BashivanRYC15}, employing Support Vector Machines (SVM), Random Forests and Sparse Logistic Regression for learning EEG representation, cannot be employed as baseline since they do not operate on whole brain signals (but on feature vectors) and are applied to other tasks (e.g., music classification, seizure detection, etc.) than visual object--evoked EEG data.\\
Table~\ref{tab:results_rnn} reports the achieved performance by the three encoder configurations with various architecture details. We also tested more complex models (e.g., using 256 nodes) but these ended up with overfitting. The classifier used to compute the accuracy is the one jointly trained in the encoder; we will use the same classifier (without any further training) also for automated visual classification on CNN-regressed EEG features.
The proposed RNN-based approach was able to reach about 83\% classification accuracy, which greatly outperforms the performance achieved by~\cite{Kaneshiro2015}, which was 29\% over 12 classes of their dataset, and 13\% on our dataset.\\
To further contribute to the research on how visual scenes are processed by the human brain, we investigated how image visualization times may affect classification performance. 
Thus far, it has been known that feature extraction for object recognition in humans happens during the first 50-120 ms~\cite{heckenlively2006principles} (stimuli propagation time from the eye to the visual cortex), whereas less is known after 120 ms. Since in our experiments, we displayed each image for 500 ms; we evaluated classification performance in different visualization time ranges, i.e., [40-480 ms], [40-160 ms], [40-320 ms] and [320-480 ms].  Table~\ref{tab:results_visualization_times} shows the achieved accuracies when using the RNN model which obtained the highest validation accuracy (see Table~\ref{tab:results_rnn}), i.e., the common 128-neuron LSTM followed by the 128-neuron output layer. Contrary to what was expected, the best performance was obtained in the time range [320-480 ms], instead of during the first 120 ms. This suggests that a key role in visual classification may be played by neural processes outside the visual cortex that are activated after initial visual recognition and might be responsible for the conception part mentioned in the introduction. Of course, this needs further and deeper investigation that are outside the scope of this paper.

\begin{table}[]
	\centering
	\begin{tabular}{ccc}
		\toprule
		\textbf{Visualization time}                 & \textbf{Max VA} & \textbf{TA at max VA} \\
		\midrule
		40-480 ms & 85.4\% & 82.9\% \\
		40-160 ms & 81.4\% & 77.5\% \\
		40-320 ms & 82.6\% & 79.7\% \\
		\textbf{320-480 ms} & \textbf{86.9\%} & \textbf{84.0\%} \\
		\bottomrule
	\end{tabular}
	\caption{Classification accuracy achieved by the RNN encoder using different portions of EEG signal data. Best results in bold.}
	\label{tab:results_visualization_times}
\end{table}

\begin{table*}
	\centering
	\tabcolsep=0.08cm 
	\begin{tabular}{cccccccccccccccccc}
		\toprule
		\multirow{2}{*}{\textbf{Feature set}} & \multirow{2}{*}{\textbf{AlexNet FT}} & \multicolumn{3}{c}{\textbf{AlexNet FE}} & \multicolumn{3}{c}{\textbf{GoogleNet}} & \multicolumn{3}{c}{\textbf{VGG}} \\
		\cmidrule(lr){3-5} \cmidrule(lr){6-8} \cmidrule(lr){9-11}
		& & \textbf{k-NN} & \textbf{Ridge} & \textbf{RF} & \textbf{k-NN} &  \textbf{Ridge} & \textbf{RF} & \textbf{k-NN} &  \textbf{Ridge} & \textbf{RF} \\
		\cmidrule(lr){1-1} \cmidrule(lr){2-2} \cmidrule(lr){3-5} \cmidrule(lr){6-8} \cmidrule(lr){9-11}		
		Average & 1.86  & 1.64	  &  1.53    & 1.52     &\underline{{\bf 0.62}} & 1.88 & 0.93 & 0.73 & 1.53 & 0.94\\
		Best 	& 2.12 & 1.94 & 1.62 & 1.56 & 3.54 & 7.06 & 4.01 &3.26  & 7.63 & 4.45\\
		\bottomrule
	\end{tabular}
	\caption{Mean square error (MSE) values obtained by different regression methods for extracting EEG features from images. ``FT'': fine-tuned; ``FE'': feature extractor. Best performance underlined and in bold.}
	\label{tab:results_regression}
\end{table*}

\subsection{CNN-based regression}
CNN-based regression aims at projecting visual images onto the learned EEG manifold. According to the results shown in the previous section, the best encoding performance is obtained given by the common 128-neuron LSTM followed by the 128-neuron output layer. This implies that our regressor takes as input single images and provides as output a 128-feature vector, which should ideally resemble the one learned by the encoder.\\
To test the regressor's performance, we used the same ImageNet subset and the same image splits employed for the RNN encoder. However, unlike the encoder's training stage, where different subjects generated different EEG signal tracks even when looking at the same image, for CNN-based regression we require that each image be associated to only one EEG feature vector, in order to avoid ``confusing'' the network by providing different target outputs for the same input. We tested two different approaches for selecting the single feature vector associated to each image:
\begin{itemize}
 \item \textit{average}: the EEG feature vector associated to an image is computed as the average over all subjects when viewing that image.
 \item \textit{best features}: for each image, the associated EEG feature vector is the one having the smallest classification loss over all subjects during RNN encoder training.
\end{itemize}

Table~\ref{tab:results_regression} shows the mean square error (MSE) obtained with each of the tested regression approaches. The lowest-error configuration, i.e., feature extraction with GoogleNet combined to k-NN regressor, was finally employed as EEG feature extractor from arbitrary images. Note that the accuracy values for \textit{average} are markedly better than the \textit{best features}' one. This is in line with the literature on cognitive neuroscience, for which changes in EEG signals elicited by visual object stimuli are typically observed when averaging data from multiple trials and subjects~\cite{Stewart2014}.

\subsection{Automated visual classification}

This section aims at demonstrating that the initial claim, i.e., that human visual capabilities can be learned and transferred to machines by testing an automated visual classifier that extracts EEG features from images (through the combination of CNN-based feature regressor - GoogleNet features with k-NN regressor according to Table~\ref{tab:results_regression}) and then classifies feature vectors using the softmax classifier trained during EEG manifold learning.  

We evaluated image classification performance on the images from our dataset's test split, which were never used in either EEG manifold learning or CNN-based feature regression, obtaining a mean classification accuracy of 89.7\%, which, albeit slightly lower than state-of-the-art CNN performance\footnote{http://image-net.org/challenges/LSVRC/2015/results}, demonstrates the effectiveness of our approach.

In order to test the generalization capabilities of our brain-learned features, we also performed an evaluation of the proposed method as a feature extraction technique, and compared it to VGG and GoogleNet (we did not test AlexNet given its lower performance as shown in Table~\ref{tab:results_regression}) as feature extractors. We tested the three (off-the-shelf) deep networks on a 30-class subset of Caltech-101 \cite{1597116} (chosen so as to avoid overlap with the classes used for developing our model) by training separate multiclass SVM classifiers (one for each network) and comparing the classification accuracy. The results are reported in Table~\ref{tab:accuracy}.\\
Our approach achieves comparable performance to GoogleNet and much better performance than VGG,  which is actually an impressive result, considering that our EEG encoder and regressor were trained on a feature space not even directly related to visual features.

\begin{table}[h!]
	\centering
	\begin{tabular}{ccccc}
		\toprule
		\textbf{GoogleNet} & \textbf{VGG} & \textbf{Our method} \\
		\midrule
		92.6\% & 80.0\% & 89.7\% \\
		\bottomrule
	\end{tabular}
	\caption{Classification accuracy achieved when using GoogleNet, VGG and the proposed method as image feature extractors for training an SVM classifier on a subset of Caltech-101.}
	\label{tab:accuracy}
\end{table}

\section{Conclusions}
In this paper we propose the first human brain--driven automated visual classification method. It consists of two stages: 1) an RNN-based method to learn visual stimuli-evoked EEG data as well as to find a more compact and meaningful representation of such data; 2) a CNN-based approach aiming at regressing images into the learned EEG representation, thus enabling automated visual classification in a ``brain-based visual object manifold''. We demonstrated that both approaches show competitive performance, especially as concerns learning EEG representation of object classes.
The promising results achieved in this first work make us hope that human brain processes involved in visual recognition can be effectively decoded for further inclusion into automated methods. Under this scenario, this work can be seen as a significant step towards interdisciplinary research across computer vision, machine learning and cognitive neuroscience for transferring human visual (and not only) capabilities to machines. It also lays the foundations for a paradigm shift in computer vision: from performance-based one to human-base computation one.

As future work, we plan \textit{a}) to develop more complex deep learning approaches for distinguishing brain signals generated from a larger number of image classes, and \textit{b}) to interpret/decode EEG-learned features in order to identify brain activation areas, band frequencies, and other relevant information necessary to uncover human neural underpinnings involved in the visual classification.

\subsubsection*{Acknowledgments}

We gratefully acknowledge the support of NVIDIA Corporation with the donation of the Titan X Maxwell GPU used for this research. We also acknowledge Dr. Martina Platania for carrying out EEG data acquisition as well as Dr. Riccardo Ricceri for supporting experimental protocol setup.

\small
\bibliographystyle{abbrv}

\begin{thebibliography}{10}
	
	\bibitem{BashivanRYC15}
	P.~Bashivan, I.~Rish, M.~Yeasin, and N.~Codella.
	\newblock Learning representations from {EEG} with deep recurrent-convolutional
	neural networks.
	\newblock In {\em To appear on ICLR 2016}, 2016.
	
	\bibitem{Shamlo2008}
	N.~Bigdely-Shamlo, A.~Vankov, R.~R. Ramirez, and S.~Makeig.
	\newblock {Brain activity-based image classification from rapid serial visual
		presentation.}
	\newblock {\em IEEE transactions on neural systems and rehabilitation
		engineering : a publication of the IEEE Engineering in Medicine and Biology
		Society}, 16(5):432--441, 2008.
	
	\bibitem{pmid23908380}
	T.~Carlson, D.~A. Tovar, A.~Alink, and N.~Kriegeskorte.
	\newblock {{R}epresentational dynamics of object vision: the first 1000 ms}.
	\newblock {\em Journal of Vision}, 13(10), 2013.
	
	\bibitem{pmid21920851}
	T.~A. Carlson, H.~Hogendoorn, R.~Kanai, J.~Mesik, and J.~Turret.
	\newblock {{H}igh temporal resolution decoding of object position and
		category}.
	\newblock {\em Journal of Vision}, 11(10), 2011.
	
	\bibitem{5492691}
	H.~Cecotti and A.~Graser.
	\newblock Convolutional neural networks for p300 detection with application to
	brain-computer interfaces.
	\newblock {\em IEEE Transactions on Pattern Analysis and Machine Intelligence},
	33(3):433--445, March 2011.
	
	\bibitem{pmid20302949}
	K.~Das, B.~Giesbrecht, and M.~P. Eckstein.
	\newblock {{P}redicting variations of perceptual performance across individuals
		from neural activity using pattern classifiers}.
	\newblock {\em Neuroimage}, 51(4):1425--1437, Jul 2010.
	
	\bibitem{1597116}
	L.~Fei-Fei, R.~Fergus, and P.~Perona.
	\newblock One-shot learning of object categories.
	\newblock {\em IEEE Transactions on Pattern Analysis and Machine Intelligence},
	28(4):594--611, April 2006.
	
	\bibitem{heckenlively2006principles}
	J.~R. Heckenlively and G.~B. Arden.
	\newblock {\em Principles and practice of clinical electrophysiology of
		vision}.
	\newblock MIT press, 2006.
	
	\bibitem{Hochreiter:1997:LSM:1246443.1246450}
	S.~Hochreiter and J.~Schmidhuber.
	\newblock Long short-term memory.
	\newblock {\em Neural Comput.}, 9(8):1735--1780, 1997.
	
	\bibitem{Kaneshiro2015}
	B.~Kaneshiro, M.~{Perreau Guimaraes}, H.-S. Kim, A.~M. Norcia, and P.~Suppes.
	\newblock {A Representational Similarity Analysis of the Dynamics of Object
		Processing Using Single-Trial EEG Classification}.
	\newblock {\em Plos One}, 10(8):e0135697, 2015.
	
	\bibitem{Kapoor2008}
	A.~Kapoor, P.~Shenoy, and D.~Tan.
	\newblock {Combining brain computer interfaces with vision for object
		categorization}.
	\newblock {\em 26th IEEE Conference on Computer Vision and Pattern Recognition,
		CVPR}, 2008.
	
	\bibitem{pmid10777794}
	Z.~Kourtzi and N.~Kanwisher.
	\newblock {{C}ortical regions involved in perceiving object shape}.
	\newblock {\em J. Neurosci.}, 20(9):3310--3318, May 2000.
	
	\bibitem{krizhevsky2012imagenet}
	A.~Krizhevsky, I.~Sutskever, and G.~E. Hinton.
	\newblock Imagenet classification with deep convolutional neural networks.
	\newblock In {\em Advances in neural information processing systems}, pages
	1097--1105, 2012.
	
	\bibitem{pmid19837629}
	P.~Mirowski, D.~Madhavan, Y.~Lecun, and R.~Kuzniecky.
	\newblock {{C}lassification of patterns of {E}{E}{G} synchronization for
		seizure prediction}.
	\newblock {\em Clin Neurophysiol}, 120(11):1927--1940, Nov 2009.
	
	\bibitem{niedermeyer2005electroencephalography}
	E.~Niedermeyer and F.~L. da~Silva.
	\newblock {\em Electroencephalography: basic principles, clinical applications,
		and related fields}.
	\newblock Lippincott Williams \& Wilkins, 2005.
	
	\bibitem{pmid18829969}
	H.~P. Op~de Beeck, K.~Torfs, and J.~Wagemans.
	\newblock {{P}erceived shape similarity among unfamiliar objects and the
		organization of the human object vision pathway}.
	\newblock {\em J. Neurosci.}, 28(40):10111--10123, Oct 2008.
	
	\bibitem{pmid17643089}
	M.~V. Peelen and P.~E. Downing.
	\newblock {{T}he neural basis of visual body perception}.
	\newblock {\em Nat. Rev. Neurosci.}, 8(8):636--648, Aug 2007.
	
	\bibitem{ILSVRC15}
	O.~Russakovsky, J.~Deng, H.~Su, J.~Krause, S.~Satheesh, S.~Ma, Z.~Huang,
	A.~Karpathy, A.~Khosla, M.~Bernstein, A.~C. Berg, and L.~Fei-Fei.
	\newblock {ImageNet Large Scale Visual Recognition Challenge}.
	\newblock {\em International Journal of Computer Vision (IJCV)},
	115(3):211--252, 2015.
	
	\bibitem{export:64271}
	P.~Shenoy and D.~Tan.
	\newblock Human-aided computing: Utilizing implicit human processing to
	classify images.
	\newblock In {\em CHI 2008 Conference on Human Factors in Computing Systems},
	2008.
	
	\bibitem{Simanova2010}
	I.~Simanova, M.~van Gerven, R.~Oostenveld, and P.~Hagoort.
	\newblock {Identifying object categories from event-related EEG: Toward
		decoding of conceptual representations}.
	\newblock {\em PLoS ONE}, 5(12), 2010.
	
	\bibitem{simonyan2014very}
	K.~Simonyan and A.~Zisserman.
	\newblock Very deep convolutional networks for large-scale image recognition.
	\newblock {\em arXiv preprint arXiv:1409.1556}, 2014.
	
	\bibitem{Stewart2014}
	A.~X. Stewart, A.~Nuthmann, and G.~Sanguinetti.
	\newblock {Single-trial classification of EEG in a visual object task using ICA
		and machine learning}.
	\newblock {\em Journal of Neuroscience Methods}, 228:1--14, 2014.
	
	\bibitem{stober}
	S.~Stober, A.~Sternin, A.~M. Owen, and J.~A. Grahn.
	\newblock Deep feature learning for {EEG} recordings.
	\newblock In {\em To appear on ICLR 2016}, 2016.
	
	\bibitem{Subasi2010}
	A.~Subasi and M.~Ismail~Gursoy.
	\newblock {EEG} signal classification using {PCA}, {ICA}, {LDA} and {S}upport
	{V}ector {M}achines.
	\newblock {\em Expert Syst. Appl.}, 37(12):8659--8666, Dec. 2010.
	
	\bibitem{szegedy2015going}
	C.~Szegedy, W.~Liu, Y.~Jia, P.~Sermanet, S.~Reed, D.~Anguelov, D.~Erhan,
	V.~Vanhoucke, and A.~Rabinovich.
	\newblock Going deeper with convolutions.
	\newblock In {\em Proceedings of the IEEE Conference on Computer Vision and
		Pattern Recognition}, pages 1--9, 2015.
	
	\bibitem{pmid22983495}
	C.~Wang, S.~Xiong, X.~Hu, L.~Yao, and J.~Zhang.
	\newblock {{C}ombining features from {E}{R}{P} components in single-trial
		{E}{E}{G} for discriminating four-category visual objects}.
	\newblock {\em J Neural Eng}, 9(5):056013, Oct 2012.
	
\end{thebibliography}

\end{document}